\documentclass[conference]{IEEEtran}
\IEEEoverridecommandlockouts
\usepackage{cite}
\usepackage{amsmath,amssymb,amsfonts}
\usepackage{algorithmic}
\usepackage{graphicx}
\usepackage{textcomp}
\usepackage{xcolor}
\usepackage{eso-pic}
\usepackage{booktabs}
\usepackage{comment}
\usepackage{multirow}
\usepackage{tabularray}

\usepackage{hyperref}  
\usepackage{xcolor}    

\hypersetup{
    colorlinks=true,        
    linkcolor=blue,         
    citecolor=blue,          
    urlcolor=magenta        
}

\def\BibTeX{{\rm B\kern-.05em{\sc i\kern-.025em b}\kern-.08em
    T\kern-.1667em\lower.7ex\hbox{E}\kern-.125emX}}

\newcommand{\conf}[1]{
\AddToShipoutPictureBG*{
\AtPageUpperMyright{#1}
}
}
\makeatletter
\def\ps@IEEEtitlepagestyle{%
    \def\@oddfoot{\mycopyrightnotice}%
    \def\@evenfoot{}%
}

\def\mycopyrightnotice{
  {\footnotesize 979-8-3315-5542-9/25/\$31.00 ~\copyright2025 ~IEEE\hfill} 
  \gdef\mycopyrightnotice{}
}

\let\old@ps@IEEEtitlepagestyle\ps@IEEEtitlepagestyle
\def\confheader#1{%
    \def\ps@IEEEtitlepagestyle{%
        \old@ps@IEEEtitlepagestyle%
        \def\@oddhead{\strut\hfill#1\hfill\strut}%
        \def\@evenhead{\strut\hfill#1\hfill\strut}%
    }%
    \ps@headings%
}
\makeatother

\newcommand\AtPageUpperMyright[1]{\AtPageUpperLeft{
 \put(\LenToUnit{0.28\paperwidth},\LenToUnit{-1cm}){
     \parbox{0.78\textwidth}{\raggedleft\fontsize{9}{11}\selectfont #1}}
 }}

\title{Quantum Machine Learning for Image Classification: A Hybrid Model of Residual Network with Quantum Support Vector Machine\\
}

\conf{ Next-Generation Computing, IoT and Machine Learning (NCIM 2025)}

\author{
\IEEEauthorblockN{
Md. Farhan Shahriyar$^{*}$, 
Gazi Tanbhir$^{\dagger}$, 
Abdullah Md Raihan Chy$^{\dagger}$}

\IEEEauthorblockA{
$^{*}$\textit{Department of Computer Science and Engineering}\\ 
{Hajee Mohammad Danesh Science and Technology University} \\
$^{*}$$^{\dagger}$\textit{R \& D, ZogBiyog, Dhaka, Bangladesh} \\
}
}

\begin{document}
\maketitle

\begin{abstract}
Recently, there has been growing attention on combining quantum machine learning (QML) with classical deep learning approaches as computational techniques are key to improving the performance of image classification tasks.This study presents a hybrid approach that uses ResNet-50 (Residual
Network) for feature extraction and Quantum Support Vector Machines (QSVM) for classification in the context of potato disease detection. Classical machine learning as well as deep learning models often struggle with high-dimensional and complex datasets necessitating advanced techniques like quantum computing to improve classification efficiency.

In our research, we use ResNet-50 to extract deep feature representations from RGB images of potato diseases. These features are then subjected to dimensionality reduction using  PCA (Principal Component Analysis) . The resulting features are processed through QSVM models which apply various quantum feature maps—such as ZZ, Z and Pauli-X to transform classical data into quantum states. To assess the model's performance we compared it with classical ML such as Support Vector Machine (SVM) and Random Forest (RF) using 5-fold stratified cross-validation for a comprehensive evaluation. The experimental results demonstrate that the Z-feature map-based QSVM outperforms classical models achieving an accuracy of 99.23\% surpassing both SVM and RF models.

This research highlights the advantages of integrating quantum computing into image classification and also provides disease detection solution into the potential of hybrid quantum-classical model.

\end{abstract}


\section{Introduction}
In today's data-centric world the efficient use of data greatly enhances problem-solving capabilities and drives the creation of intelligent systems within society. Various forms of data have distinct applications and among them image data play a crucial role in numerous domains such as healthcare diagnostics, plant life studies, and many more \cite{q1},\cite{q2},\cite{q3}. Proper handling of image data presents challenges, particularly in classification tasks which are fundamental to computer vision.

Image classification has progressed from conventional techniques to machine learning-driven methods over time. The introduction of machine learning has greatly enhanced both the accuracy and interpretability of image classification tasks. While conventional machine learning models such as Random Forest, K-Nearest Neighbors, Naïve Bayes and Decision Trees, have demonstrated effectiveness in specific scenarios, they often struggle with high-dimensional and complex image data. In contrast, deep learning models, particularly Convolutional Neural Networks (CNNs), have transformed image classification with architectures like AlexNet \cite{q4}, ResNet\cite{q5}, MobileNet \cite{q6}, and GoogleNet\cite{q7}. Despite their success, these models face challenges due to high dimensionality and intricate patterns in image data. Consequently, quantum machine learning (QML) emerges as a promising alternative using quantum parallelism to handle complex computations efficiently. As computational paradigms shift from classical to quantum machine learning hybrid approaches that integrate both methodologies require deeper investigation for improved image classification performance.

\subsection{Classical Machine Learning for Image Classification}

Classical machine learning image classification involves representing images as multi-dimensional arrays of pixel values, which are then processed through feature extraction techniques like edge detection (e.g., Sobel operator) or descriptors such as HOG or SIFT\cite{q8}. These methods transform the raw image data into a set of meaningful features. Dimensionality reduction often using PCA is then applied to reduce the feature space's complexity\cite{q9}. The extracted features are fed into classification algorithms like SVM which maximize the margin between classes by solving an optimization problem or k-NN which classifies based on the majority class of nearby data points using distance metrics like Euclidean distance. The model is trained on labeled data to minimize classification error and once trained, it can predict the class of new, unseen images. The computational steps rely on convolution, optimization, and distance-based techniques to transform image data into actionable classifications.Image classification algorithms can be categorized based on their architecture. Traditional machine learning methods like SVM and k-NN rely on manually extracted features. CNNs (e.g., AlexNet, VGGNet, ResNet) automatically learn hierarchical features from raw images\cite{q10}. Efficient architectures like MobileNet and EfficientNet are designed for mobile and resource-constrained devices, optimizing computational efficiency. Transformer-based models like Vision Transformer (ViT) and DeiT use attention mechanisms for improved performance\cite{q11}. Additionally, Generative models such as GANs and VAEs focus on learning latent representations, which can also be applied for image classification tasks.

\subsection{Quantum Machine Learning for Image Classification}

In Quantum Machine Learning (QML) for image classification, principles of quantum computing like superposition, entanglement and interference are applied to enhance the efficiency of image data processing. Image data traditionally represented as vectors or matrices is encoded into quantum states using encoding methods like amplitude encoding where classical image features are mapped into quantum bits (qubits). Since qubits represent binary values image data must be normalized to values in the range of 0 to 1 or -1 to 1 to be effectively mapped onto quantum states. This scaling ensures that classical data fits within the constraints of quantum states and can be processed by its circuits. Quantum circuits combination of quantum gates (e.g., Hadamard, CNOT, and Pauli gates) manipulate these qubits to extract meaningful features. Quantum operations exploit the parallelism inherent in superposition allowing quantum classifiers to handle multiple possibilities simultaneously. After quantum features are processed upon measurement a quantum state collapses to one of its basis states providing the predicted class label. Model parameter optimization is performed through quantum algorithms like Quantum Gradient Descent or Variational Quantum Eigensolver (VQE), which update model weights to minimize the loss function\cite{q12},\cite{q13}. These quantum algorithms can provide improvements in efficiency and performance over classical approaches under certain conditions.

In quantum image classification, algorithms are categorized into quantum-enhanced architectures, such as Quantum Convolutional Neural Networks (QCNNs) which use quantum gates for feature extraction and QSVMs ( Quantum Support Vector Machines)  which apply quantum kernels for feature processing. Other models include QNNs which combine quantum states and classical learning and Quantum Generative Models (e.g., GANs and VAEs) which generate and classify image data using quantum circuits offering new advantages in processing complex data\cite{q14}.

\subsection{Classical-Quantum Hybrid Approaches for Image Classification}

Image classification using classical-quantum machine learning (QML) has attracted considerable attention due to its potential to enhance image processing tasks\cite{q15}. Classical techniques such as CNNs or other deep learning architectures excel in feature extraction while quantum computing addresses optimization challenges through quantum algorithms like entanglement and superposition. By combining these methods, image processing can be accelerated, and classification performance improved. Quantum models such as QNNs and quantum support vector machines offer faster data handling and better outcomes\cite{q16}. However, despite the development of numerous QML models including quantum ansatz and parameterized quantum circuits further research is necessary for broader applicability. One of the main limitations of current quantum hardware is the limits number of qubits which hinders the practical deployment of quantum models for its more generalized applications. Additionally, the loss and cost functions in hybrid models are critical for optimizing the balance between classical and quantum components ensuring efficient convergence. While QML holds great promise particularly for managing dynamic features and complex image classification tasks through hybrid models ongoing research is required to fully realize the potential of quantum algorithms and hardware. In the meantime, a hybrid approach in which classical deep learning models perform feature extraction before passing data to quantum models ( different CNN or deep learning based models feature extraction \cite{q17},\cite{q18},\cite{q19},\cite{q20} ) can improve performance. This strategy encourages further exploration of classical-quantum integrations and supports the development of more advanced QML methods for large-scale image classification

To address the challenges in potato disease prediction and enhance the accuracy of detection, our contribution in this research highlights:

\begin{itemize}
    \item Hybrid ResNet-50 feature extraction and QSVM Z-feature map model for potato disease prediction which outperforms classical machine learning algorithms ensuring more accurate and efficient disease detection.
\end{itemize}

\begin{itemize}
    \item Utilization of advanced deep learning techniques in conjunction with quantum support vector machines (QSVM) to improve the accuracy of disease classification addressing challenges in classical machine learning systems.
\end{itemize}

The rest of the study is organized as follows: Section \ref{sec:background} presents the background study providing an overview of existing methods and challenges in potato disease prediction. Section \ref{sec:methodology} details the methodology including the hybrid ResNet-50 and QSVM Z-feature map model used for disease classification. The Results and Analysis section \ref{sec:result} showcases the evaluation of the proposed model by comparing it with classical approaches. Finally, the Conclusion and Future Work section \ref{sec:conclusion} summarizes the findings and outlines potential directions for further research.

\section{Background Study}
\label{sec:background}

This study \cite{q21} utilizes quantum support vector machines (QSVM) for classifying brain tumors as malignant or benign using the Brats 2015 dataset. The QSVM model tested on both quantum simulators and real-time machines was 188 times faster and 1.60\% more accurate (95\%) than its classical counterpart on a 32-qubit simulator. On a 5-qubit superconducting processor, QSVM was 24.19\% faster while maintaining the same accuracy highlighting the superiority of quantum models in speed and performance.

An ensemble quantum machine learning model \cite{q22} was proposed to classify Alzheimer’s disease using merged datasets from the Alzheimer’s Disease Neuroimaging Initiative (ADNI). Features were extracted with VGG16 and ResNet50 models and classified by the quantum model.The proposed model achieved an impressive F1-score with a low error rate of 1.63\% outperforming existing methods in Alzheimer's disease detection.

A hybrid quantum-classical algorithm (HQCA) \cite{q23} for image classification was introduced using a quantum kernel for improved efficiency in Support Vector Classification. Amplitude encoding was applied to image data, and features were stored using the ZZFeatureMap in the Qiskit framework. This approach demonstrates the potential of hybrid quantum-classical systems to enhance image classification tasks.

This study \cite{q24} uses a quantum support vector machine  in neonatal retinal fundus images to classify Retinopathy of Prematurity . High-level features extracted from segmented retinal vessels using SIFT and SURF were classified by the QSVM model. Compared to classical models like ResNet50 and SVM, QSVM achieved the highest accuracy (86.7\%) in detecting ROP, showing its potential for effective retinal disease diagnosis.

This study \cite{q25} proposes a QSVC model with grey scale images for binary classification. To enhance the prediction rate and tackle the noise from Noisy Intermediate-Scale Quantum (NISQ) systems pre-processing techniques like feature selection as well as state preparation were applied. The model was tested on datasets like Extended Gun, Knife, CohnKanade  and FER2013. The QSVC achieved higher accuracy compared to the classical SVM, with scores of 98\%, 98\%, 93\%, and 92\%, respectively, compared to the classical SVM's accuracy of 95\%, 83\%, 87\%, and 83\%.

This work \cite{q26} combines quantum computing and conventional machine learning for skin disease classification using the HAM10000 dataset. Quantum libraries like PennyLane and Qiskit were employed, and various qubit rotation encodings using Pauli X, Y, and Z gates were explored in the proposed Quanvolutional neural network. Features extracted by MobileNet were used to build a Quantum Support Vector Classifier. Among the models tested, the Quanvolutional network with RY qubit rotation and PauliZ gate achieved the highest classification accuracy of 82.86\%, surpassing other models. The Quantum Support Vector Classifier however achieved 72.5\% comparable to pre-trained models like ResNet50 and DenseNet.

This study \cite{q27} proposes a hybrid approach combining classical feature extraction with quantum machine learning to address resource demands and noise issues. A ResNet10-inspired convolutional autoencoder is used to reduce the dimensionality of the dataset and extract meaningful features before feeding them into a quantum-enhanced QSVM. The QSVM is chosen for its ability to work with small datasets and its short-depth quantum circuits which help mitigate noise. The autoencoder is trained to minimize mean squared error in image reconstruction. The approach was tested on HTRU-1, MNIST, and CIFAR-10 datasets with a quantum one-class support vector machine (QOCSVM) applied to the imbalanced HTRU-1 dataset. The autoencoder performed well on MNIST with high classification accuracy while CIFAR-10 showed reduced performance due to image complexity, and HTRU-1 struggled due to its imbalance. The results emphasize the importance of balancing classical feature extraction and quantum methods for optimal prediction performance.

\section{Methodology}
\label{sec:methodology}

\begin{figure}[htbp] 
    \centering
    \includegraphics[width=1\linewidth]{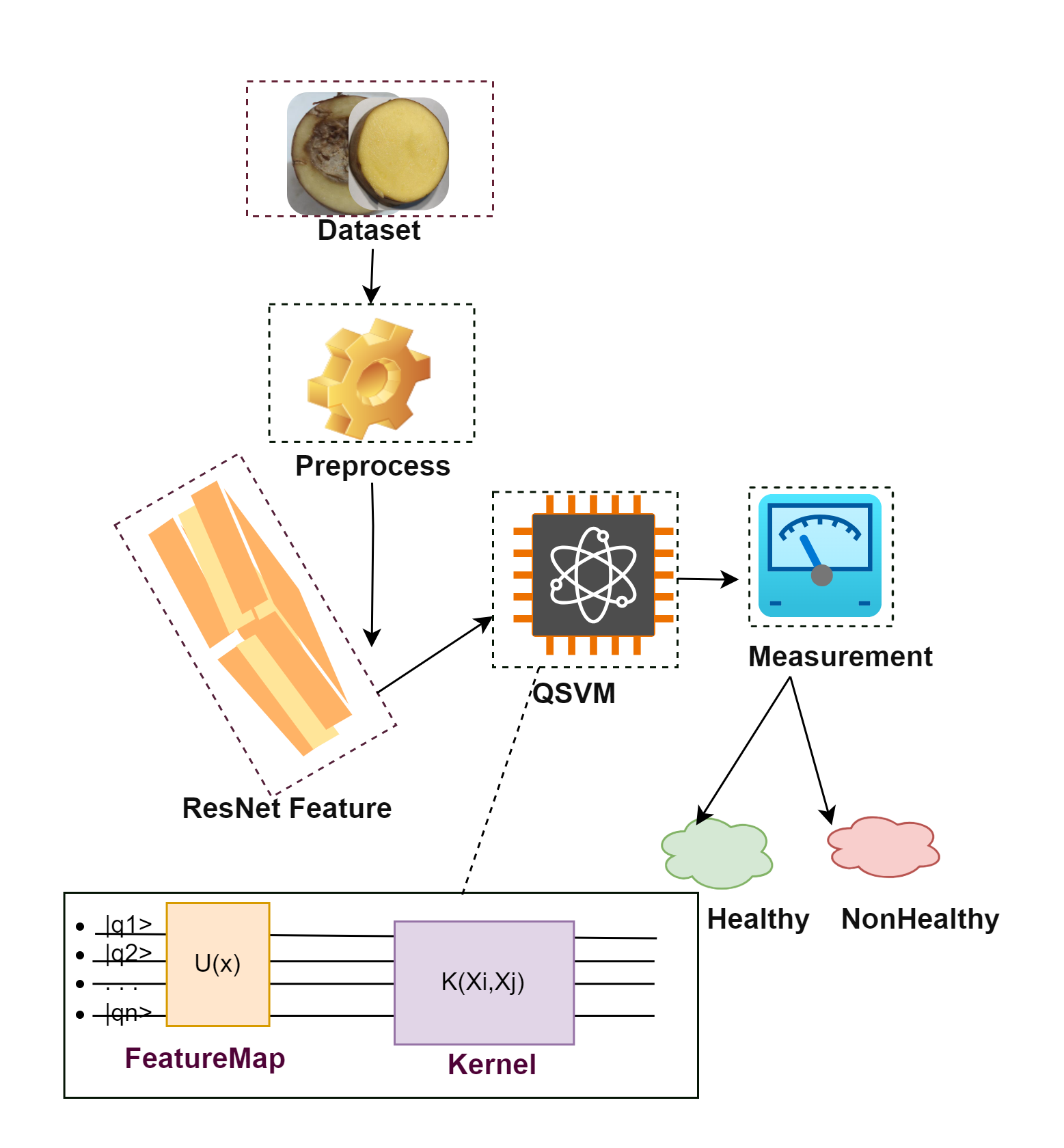}
    \caption{Classical-Quantum ResNet QSVM Model for Disease Classification}
    \label{fig:resnetqsvm}
\end{figure}

This section outlines the methodology applied for the classification of potato disease images with a primary focus on the integration of both classical machine learning models and quantum machine learning techniques. The methodology consists of several key stages: dataset extraction, data preprocessing, feature extraction using ResNet50, dimensionality reduction through Principal Component Analysis (PCA) classification using classical models such as SVM and Random Forest and the incorporation of various quantum feature maps (Z, ZZ, PauliX) with quantum kernels to potentially enhance classification performance.These steps are illustrated in detail in Figure \ref{fig:resnetqsvm}, providing a comprehensive view of the process from start to finish.The QSVM's feature map circuit and kernel utilization is also shown here.

\begin{figure}[htbp] 
    \centering
    \includegraphics[width=1\linewidth]{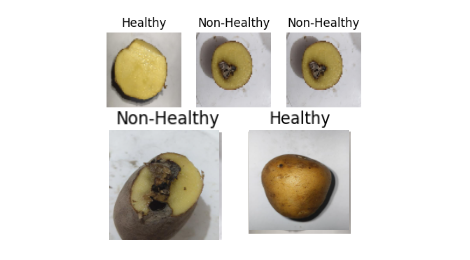}
    \caption{Potato Images of Healthy and Non-Healthy (Diseased) }
    \label{fig:potatodiseasesphoto}
\end{figure}

\subsection{Dataset Description and Extraction} 
This study utilizes a comprehensive dataset containing five potato classifications: Healthy Potato, Blackspot Bruising , Soft Rot , Brown Rot , and Dry Rot . The dataset \cite{q28} comprises 495 original potato images sourced from a larger collection of potato disease images available on Mendeley. The focus of this study is on two specific classes: Healthy Potato and Soft Rot Disease. The objective is to develop quantum machine learning-based algorithms capable of accurately distinguishing between healthy potatoes and those affected by Soft Rot Disease as nonhealthy photos in figure \ref{fig:potatodiseasesphoto}. This approach aims to enhance the detection and classification precision of potato diseases using quantum machine learning techniques.

\subsection{Image Preprocessing and Feature Extraction Using ResNet50}

The first critical step in preparing the images for classification is preprocessing. The images are resized to 224x224 pixels the standard input size required by the ResNet50 model.It is a pre-trained convolutional neural network that was originally trained on the ImageNet dataset. After resizing, each image was converted into a NumPy array using a Keras function. Additionally, the images underwent normalization using the \texttt{preprocess\_input}  function, which ensures that the pixel values are standardized according to the specific preprocessing requirements of ResNet50.

ResNet50 is employed in its feature extraction mode, meaning that the top layer (usually used for classification) was omitted. Instead, the model outputs feature vectors from the last convolutional block which were further processed for classification. The ResNet50 model outputs a 2048-dimensional feature vector for each image after performing average pooling. These feature vectors were flattened into 1D arrays to serve as input for further machine learning algorithms.

\subsection{Dimensionality Reduction with Principal Component Analysis (PCA)} 
Given the high-dimensional nature of the features extracted from ResNet50, dimensionality reduction was performed to simplify the feature space and reduce the risk of overfitting. PCA was chosen for this purpose. PCA is an unsupervised method that maps data to a lower-dimensional space preserving the maximum variance. This step is essential for improving model efficiency and reducing computational cost.

In this study, three different configurations of PCA were tested, with 3, 6, and 9 components. The first step in applying PCA was to fit the training data to the PCA model, followed by transforming both the training and testing datasets. PCA reduced the dimensionality of the extracted feature vectors while preserving as much information as possible. After PCA transformation, the feature vectors were scaled using MinMaxScaler to normalize the values to the range of [-1, 1] making them more suitable for machine learning models.

\subsection{Model Selection and Training (Classical Models)} 
The focus of this study is on classifying potato disease images using ResNet extracted features combined with two classical machine learning models: Random Forest and Support Vector Machine (SVM) . These models were selected for their strong classification performance and their capability to handle high-dimensional data.

The SVM classifier employed a radial basis function (RBF) kernel with a regularization parameter. This model seeks to identify the optimal hyperplane that best separates the two classes (Healthy and Nonhealthy) in the transformed feature space.

The Random Forest classifier was set up with default estimators (trees) each trained on a data subset using bootstrapping. RF as an ensemble learning of multiple decision trees combines their predictions to enhance classification accuracy and robustness.

Both models were trained on feature vectors reduced by PCA and evaluated through Stratified K-Fold cross-validation. This method ensures that each fold preserves the class distribution from the original dataset leading to a more balanced and reliable performance evaluation.

\subsection{Stratified K-Fold Cross-Validation} 
To ensure a reliable evaluation of model performance Stratified K-Fold cross-validation was utilized. This technique divides the dataset into 5 folds maintaining an equal distribution of the two classes in each fold.Trained on 4 folds the model is tested on the remaining fold in each iteration repeating the process for all 5 folds which results in 5 distinct training and testing sets.

\subsection{Quantum Kernel Integration (Quantum Machine Learning)} 
In addition to the classical models, this study explored the use of quantum ML techniques specifically the integration of a quantum kernel into the classification process. Quantum ML has gained attention because of its potential to capture complex patterns in large datasets more effectively than classical methods.

For each input data point \( x \in X \), the feature map \( \varphi \) applies a quantum operation to create a quantum state:

\begin{equation}
|\varphi(x)\rangle = U_{\varphi}(x) |0\rangle
\end{equation}

where \( |0\rangle \) is the initial quantum state (e.g., all qubits set to \( |0\rangle \)) and \( U_{\varphi}(x) \) is the unitary transformation that depends on the data \( x \).

To integrate quantum machine learning into the workflow the QSVC was employed. The core idea behind the QSVC is to replace the classical kernel with a quantum kernel. In this study, the quantum kernel was constructed using a FidelityQuantumKernel, which calculates the fidelity between quantum states. The kernel was designed using a feature map, and the fidelity of quantum computations was computed using a custom quantum sampler.

Once the quantum kernel was defined, the QSVC was trained using the quantum kernel and evaluated in the same manner as the classical models. The quantum kernel was expected to improve classification performance by leveraging quantum computational advantages particularly in handling complex patterns in the data.

In QSVM different feature map circuits are used each with a distinct approach to convert classical data into quantum states .

The ZZ Feature Map induces entanglement among features which is beneficial for datasets where feature interactions are critical. This map captures the relationships between features and represents them as quantum states.

The Z Feature Map applies independent rotations for each feature, treating them separately. It’s ideal when individual features are more important than their interactions, as it doesn’t create entanglement.

The Pauli-X Feature Map rotates features around the X-axis, offering an alternative encoding. It can be beneficial for certain data distributions, providing a different way of mapping data into the quantum space. This quantum-enhanced kernel helps QSVM create better decision boundaries in complex datasets. \textbf{ZZ Feature Map} is best for capturing interactions, \textbf{Z Feature Map} works well for simple, independent features, and \textbf{Pauli-X Feature Map} provides an alternative encoding for different distributions.

\section{Results and Analysis}
\label{sec:result}

The classical-quantum hybrid ResNet Quantum Support Vector Machine potato disease classification model experiment is implemented on Google Colab using the CPU environment using both Qiskit and TensorFlow Keras libraries. Qiskit is used to incorporate quantum computing elements into the model specifically for the QSVM component while TensorFlow Keras is utilized for the classical deep learning architecture, particularly for the ResNet (Residual Network) component. The hybrid approach combines the power of classical neural networks for feature extraction and the quantum advantage of QSVM for enhanced classification. The experiment is designed to classify potato disease images, using a combination of classical and quantum techniques to improve prediction accuracy. Google Colab provides a convenient cloud-based platform for running the experiment allowing for easy access to both classical and quantum tools without the need for specialized hardware.

The performance of the proposed framework is extensively evaluated using several standard classification metrics with results compared against baseline models. This analysis provides a comprehensive examination of the findings.

The table \ref{resultqsvm} and \ref{resultcmp} shows the accuracy of different machine learning models SVM, RF and QSVM with various feature maps (ZZ, Z, and PauliX) under different numbers of PCA components (3, 6, and 9), using 5-fold cross-validation. For the classical models SVM and RF show consistent performance across different PCA components with RF achieving the highest accuracy particularly at 3 and 9 components. On the quantum side the ZZ and Z feature maps perform well with the Z feature map consistently achieving the highest accuracy (0.9923) across the various PCA components. In contrast, the PauliX feature map does not exhibit much variability maintaining an accuracy of 0.5658 across all configurations. These results, obtained with 5-fold cross-validation suggest that while classical models like RF show stable and high performance the quantum models particularly with the Z feature map are able to approach or exceed classical model performance with the right feature map choice.

\begin{table}
\centering

\caption{Accuracy of Classical ResNet50 with QSVM  Models for Different Feature Map}
\begin{tblr}{
  width = \linewidth,
  colspec = {Q[169]Q[149]Q[169]Q[149]},
}
\hline
PCA Components  & ZZ Feature Map  & Z Feature Map  & PauliX Feature Map  \\
\hline
3                           & 0.9615                & 0.9923               & 0.5658                    \\
6                            & 0.9154                & 0.9846               & 0.5658                    \\
9                 & 0.6662                & 0.9923               & 0.5658 \\ \hline 
\label{resultqsvm}
\end{tblr}

\end{table}

\begin{table}
\centering
\caption{Performance Comparison of Proposed Quantum Model with Classical Models}
\begin{tblr}{
  width = \linewidth,
  colspec = {Q[169]Q[139]Q[109]Q[119]},
}
\hline
PCA Components & SVM  & Random Forest   & \textbf{QSVM (Z Feature Map)}   \\
\hline
3              & 0.5658       & 0.9689                                & \textbf{0.9923}                                   \\
6              & 0.5658       & 0.9455                               & \textbf{0.9846}                                  \\
9              & 0.5658       & 0.9766                                & \textbf{0.9923}                \\ \hline 
\label{resultcmp}
\end{tblr}
\end{table}

\section{Conclusion and Future Work}
\label{sec:conclusion}

This research successfully executed and evaluated a hybrid classical-quantum ResNet-QSVM model for potato disease classification. The model utilizes the power of classical deep learning via the ResNet architecture for feature extraction combined with the quantum advantage of Quantum Support Vector Machine (QSVM) to enhance classification accuracy. Implemented on Google Colab with both Qiskit and TensorFlow Keras, the hybrid approach demonstrates significant potential in improving prediction accuracy for potato disease detection. Our results emphasize the superiority of the Z-feature map in QSVM achieving near-perfect accuracy (0.9923) surpassing classical machine learning models such as ResNet-Random Forest and ResNet-SVM. Furthermore, the comparison of classical models with quantum-based techniques reveals that the Z-feature map outperforms both the ZZ and Pauli X feature maps.

Although our research has demonstrated substantial success several promising directions for future work remain:

\begin{itemize}
    \item \textbf{Feature Map Optimization:} Explore the potential of alternative quantum feature maps (e.g., PauliZ) and investigate quantum computing strategies, such as the inclusion of a single quantum layer to further enhance classification performance.
    \item \textbf{Scalability with Quantum Hardware:} Assess the scalability of the hybrid model by transitioning from cloud-based simulations on Google Colab to actual quantum hardware, which could improve model accuracy by utilizing more qubits.
    \item \textbf{Hybrid Quantum-Classical Ensemble:} 
Explore the integration of ensemble methods that combine quantum and classical models aiming to enhance the stability, performance, and generalization of the potato disease classification model.
\end{itemize}

By pursuing these future research directions, we can refine and advance the capabilities of quantum-assisted AI systems that contribute to more accurate and scalable secure solutions for complex prediction tasks.

Ultimately, this will have a significant impact on agricultural disease detection and other fields that rely on data-driven decision making.

\bibliographystyle{IEEEtran}
\bibliography{references}

\end{document}